\documentclass[journal,10pt,onecolumn]{IEEEtran}
\usepackage{graphicx}
\usepackage{balance}
\usepackage{comment}
\usepackage{cite}
\usepackage{amssymb}
\usepackage[tight,footnotesize]{subfigure}
\usepackage[active]{srcltx}
\usepackage{amsmath}
\usepackage{mathtools}

\graphicspath{{./figs/}}
\usepackage{eurosym}
\usepackage[plain]{algorithm}
\usepackage{algorithmic}
\usepackage{multicol}
\usepackage{dsfont}
\usepackage{amsfonts}

\usepackage{cases}
\usepackage{xcolor}
\begin{document}

\title{Rolling Horizon Coverage Control with Collaborative Autonomous Agents}

\author{Savvas~Papaioannou,~Panayiotis~Kolios,~Theocharis~Theocharides,\\~Christos~G.~Panayiotou~ and ~Marios~M.~Polycarpou% <-this % stops a space
\thanks{The authors are with the KIOS Research and Innovation Centre of Excellence (KIOS CoE) and the Department of Electrical and Computer Engineering, University of Cyprus, Nicosia, 1678, Cyprus. {\tt\small \{papaioannou.savvas, pkolios, ttheocharides, christosp, mpolycar\}@ucy.ac.cy}}
}

\markboth{Philosophical Transactions of the Royal Society A, Volume 383, Issue 2289, 2025, doi:10.1098/rsta.2024.0146}%
{Papaioannou \MakeLowercase{\textit{et al.}}: Rolling Horizon Coverage Control with Collaborative Autonomous Agents}

\maketitle
\begin{abstract}
This work proposes a coverage controller that enables an aerial team of distributed autonomous agents to collaboratively generate non-myopic coverage plans over a rolling finite horizon, aiming to cover specific points on the surface area of a 3D object of interest. The collaborative coverage problem, formulated, as a distributed model predictive control problem, optimizes the agents' motion and camera control inputs, while considering inter-agent constraints aiming at reducing work redundancy. The proposed coverage controller integrates constraints based on light-path propagation techniques to predict the parts of the object's surface that are visible with regard to the agents' future anticipated states. This work also demonstrates how complex, non-linear visibility assessment constraints can be converted into logical expressions that are embedded as binary constraints into a mixed-integer optimization framework. The proposed approach has been demonstrated through simulations and practical applications for inspecting buildings with unmanned aerial vehicles (UAVs).
\end{abstract}


\section{Introduction} \label{sec:Introduction}
The interest in swarm systems such as systems utilizing multiple autonomous unmanned aerial vehicles (UAVs) has skyrocketed over the last few decades. Rapid advancements in robotics, automation and artificial intelligence coupled with the decreasing costs of electronic components have fuelled a remarkable surge in interest towards the technologies and applications of swarming systems. 
This work addresses the challenge of coverage planning and control using multiple collaborative intelligent autonomous agents, specifically autonomous UAVs. Coverage planning \cite{tan2021comprehensive} is crucial in several application domains including search and rescue operations and critical infrastructure inspections. It is one of the essential functionalities that can notably enhance the autonomy of existing swarming systems enabling them to execute fully automated missions in the aforementioned scenarios. 
In coverage planning our objective is to design trajectories that allow a team of autonomous mobile agents to comprehensively cover a designated area or points of interest. Concurrently we aim to optimize a specific mission goal such as minimizing the mission's duration and energy consumption of the agents. 

This work introduces a coverage control framework that optimizes both the kinematic and camera control inputs of multiple UAV agents simultaneously. This approach facilitates the creation of collaborative non-myopic coverage plans for 3D objects of interest. Specifically, the contributions of this work are the following: a) We propose a coverage planning framework designed to enable an aerial team of distributed agents to efficiently cover specific points on the surface area of an object of interest in 3D. Specifically, we develop a collaborative model predictive coverage controller that orchestrates the agents guiding them to optimally select their kinematic and camera control inputs. This process facilitates the generation of complementary non-myopic coverage plans that optimize coverage; b) we demonstrate that these collaborative coverage plans can be generated by incorporating light-path propagation constraints into the coverage controller. This integration aids in identifying parts of the scene that will be visible from the agents' projected future positions, a crucial aspect of developing effective non-myopic coverage strategies; and finally c) we demonstrate how the proposed collaborative coverage controller can be realized using mixed-integer programming techniques (MIP), by transforming complex, non-linear coverage constraints into logical expressions. We showcase its performance through both qualitative and quantitative real-world and synthetic experiments.

The rest of the paper is structured as follows: Section \ref{sec:Related_Work} reviews the relevant literature on coverage planning with single and multiple agents. Section \ref{sec:system_model} formulates the problem addressed in this study. Section \ref{sec:approach} details the proposed approach followed by an evaluation in Section \ref{sec:eval}. Section \ref{sec:wf} discusses open problems and future directions in swarm systems and finally Section \ref{sec:conclusion} concludes this work.

\section{Related Work}\label{sec:Related_Work}

With consumer drones becoming widespread and affordable coverage planning research has focused recently towards UAV-based applications. For instance, \cite{Xie2020} uses a traveling-salesman approach for UAV-based coverage of 2D polygonal areas whereas the authors in \cite{li2011} use exact cellular decomposition to cover polygonal planar regions with a UAV agent equipped with a downward facing camera. 
Regarding the use of multiple UAVs for this problem, the work in \cite{chleboun2022improved} proposes an artificially weighted spanning tree algorithm for distributed coverage in planar regions involving multiple UAV agents. The authors in \cite{chen2021} propose a spatiotemporal clustering-based coverage method for multiple heterogeneous UAV agents also for planar environments. Other works such as \cite{choi2020} have focused on the aspect of energy efficiency in multi-UAV coverage planning. The approach in \cite{avellar2015multi} uses mathematical programming techniques to design coverage paths that guide a team of UAV agents to cover a specified area of interest in the minimum amount of time. A UAV-centric cooperative coverage planning methodology is introduced in \cite{apostolidis2022cooperative} utilizing the simulated annealing algorithm. This approach considers the sensing and operational capabilities of each UAV, their starting positions, and the designated no-fly zones. Additionally, \cite{collins2021} outlines a multi-UAV strategy for 2D terrain coverage focusing on reducing the overall completion time by ensuring a balanced workload distribution among the UAVs. Meanwhile, \cite{perez2016} proposes a cell decomposition algorithm for multi-UAV area coverage which employs regular hexagons to optimize the coverage process. The work in \cite{Jing2020} investigates the multi-UAV coverage problem for 3D structures of interest and proposes a centralized sampling-based heuristic approach which combines the set-covering problem and the vehicle-routing problem.  The resulting set-covering vehicle routing problem is then solved with a genetic algorithm. The work in \cite{ivic2023multi} investigates the problem of 3D coverage using multiple UAV agents framing it as an offline path planning problem. The study proposes a heuristic approach based on potential fields, solved using the finite elements method. 

Despite numerous coverage planning approaches proposed in the literature, a dominant solution for enabling autonomous multi-agent coverage planning in realistic 3D environments has yet to emerge. Current state-of-the-art methods primarily focus on 2D terrain coverage \cite{Kan2020}, neglecting the complexity of 3D objects. Moreover, they often presuppose that UAVs are equipped with fixed, uncontrollable sensors (e.g., downward-facing cameras) \cite{aminzadeh2023multi} disregarding the simultaneous optimization of the UAVs' kinematics and camera control inputs during the planning phase. This simplification notably reduces the complexity of coverage planning essentially transforming it into a path-planning problem \cite{Jones2023}. Furthermore, many of the coverage planning techniques rely on simple geometric patterns (e.g., back-and-forth, zig-zag, and spiral motions) and utilize heuristics for area coverage \cite{tan2021comprehensive,ivic2023multi} which are not optimal, do not effectively generalize to 3D environments, or require centralized controllers \cite{PapaioannouCDC2023,Jing2020}. Finally, while distributed coverage planning methods have been explored to leverage the capabilities of multiple autonomous agents, these tend to yield myopic and greedy paths \cite{elmokadem2019distributed,Sunan2017} rather than collaborative, look-ahead coverage trajectories.

\section{Problem Formulation} \label{sec:system_model}

\subsection{Agent Dynamical Model} \label{ssec:kinematic_model}

We consider a collaborative team of $N$ autonomous networked aerial agents (i.e., UAVs) represented by $n \in \{1, \ldots, N\}$ evolving within a finite 3D environment denoted as $\mathcal{E} \subset \mathbb{R}^3$. These agents, with states $\boldsymbol{x}_n=[\boldsymbol{x}^{p}_n,\boldsymbol{x}^{v}_n]^\top,\forall n$ are characterized by discrete-time dynamics composed of position $(\boldsymbol{x}^{p}_n \in \mathbb{R}^3)$ and velocity $(\boldsymbol{x}^{v}_n \in \mathbb{R}^3)$ components which are described by the discrete-time state-space model:
\begin{equation}\label{eq:kinematics}
\boldsymbol{x}_n(t+1)= \boldsymbol{A} \boldsymbol{x}_n(t) + \boldsymbol{B} \boldsymbol{u}_n(t), ~\forall t, n \in \{1, \ldots, N\},
\end{equation}
\noindent where $\boldsymbol{x}_n(t)$ represents the state of the $n_\text{th}$ agent at time step $t$, and $\boldsymbol{u}_n(t) \in \mathbb{R}^3$ signifies the control input. This input denotes the applied force to the $n_\text{th}$ agent, enabling movement in a desired direction and at a certain speed. The state transition matrix $\boldsymbol{A}$, and the input matrix $\boldsymbol{B}$ are given by: $\boldsymbol{A} = 
\begin{bmatrix}
    \boldsymbol{1}_{3\times3} & \Delta T \times \boldsymbol{1}_{3\times3}\\
    \boldsymbol{0}_{3\times3} & (1-\epsilon) \times \boldsymbol{1}_{3\times3}
   \end{bmatrix}$, and $\boldsymbol{B}=\begin{bmatrix}
    \boldsymbol{0}_{3\times3} \\
     \frac{\Delta T}{m} \times \boldsymbol{1}_{3\times3}
   \end{bmatrix}$ respectively, where $\Delta T$ is the sampling interval, $\epsilon$ is the air drag coefficient, and $m$ denotes the agent's  mass which without loss of generality in this work is assumed to be the same for all agents. Moreover, the constant matrices  $\boldsymbol{1}_{3\times3}$ and $\boldsymbol{0}_{3\times3}$ denote the 3-by-3 identity and zero matrices respectively. In this work it is assumed that all agents maintain a wireless communication link with all their peers i.e., for exchanging information, and devising collaborative coverage plans. However, this assumption can be relaxed, and the role of communication is further discussed in Sec. \ref{sec:eval}.

%The resulting coverage trajectory can be used as the reference trajectory of a low-level controller (i.e., an autopilot) \cite{Simone2020,Elkaim2015,Garcia2011,Yang2016}.
%This assumption is made here merely to simplify the platform requirements for testing and the evaluation of the proposed controller in real-world settings. However, as we discuss in Sec. \ref{sec:approach} the proposed controller can be easily formulated as a distributed system.
%We should also mention here that although Eq. \eqref{eq:kinematics} does not describes the true underline aerodynamical behavior of the robot (i.e., the UAV), it can be used  to construct the desired mission trajectory, which in turn can be tracked with the appropriate (i.e., depending on the robot type such as rotor-copter or fixed-wing UAV) low-level guidance and navigation controller (e.g., an auto-pilot) \cite{Simone2020,Elkaim2015,Garcia2011,Yang2016}.

\subsection{Agent Sensing Model} \label{ssec:sensing_model}
\begin{figure*}
	\centering
	\includegraphics[width=\textwidth]{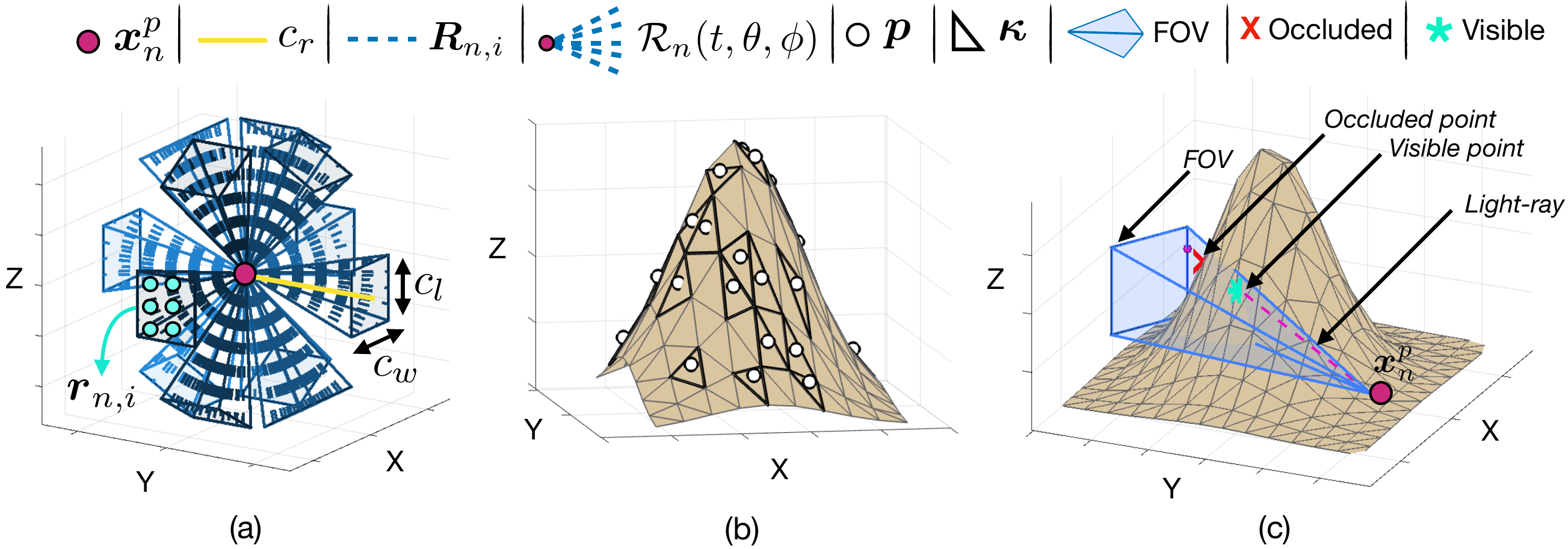}
	\caption{The figure illustrates: (a) the agent sensing model, (b) the triangulated surface area of the object of interest, and the points $\boldsymbol{p}_i$ that need to be covered, (c) the need for incorporating light-path propagation to asses the coverage of points i.e., although both points (marked with $\ast$, and $\times$) reside inside the agent's FOV, only point $\ast$ is visible as shown, this aspect is further discussed in Sec. \ref{ssec:light_path_constraints}.}
	\label{fig:fig0}
	\vspace{0mm}
\end{figure*}

We consider that each agent $n$ is outfitted with a rotating camera to monitor its environment as illustrated in Fig. \ref{fig:fig0}(a). The camera's finite field of view (FOV) is represented as a regular right pyramid, featuring four triangular sides and a rectangular base. The optical center of the camera is aligned directly above the centroid of this rectangular base. The parameters defining the camera's FOV in our model are denoted by $(c_l,c_w,c_r)$, where $c_l$ and $c_w$ correspond to the length and width of the rectangular base respectively and $c_r$ represents the pyramid's height indicating the range of the FOV. Subsequently, the camera's field of view (FOV) can be manipulated in three-dimensional space by instructing the camera controller to perform two sequential elemental rotations i.e., initially it rotates by an angle $\theta \in \Theta \subset [0,\pi)$ about the $y-$axis and then it undergoes a rotation by $\phi \in \Phi \subset [0,2\pi)$ around the $z-$axis where $\Theta$ and $\Phi$ are finite sets denoting the admissible camera rotation angles. As a result, at each time step $t$, the agent $n$ with position $\boldsymbol{x}^{p}_n(t)$ can rotate its camera's FOV through the subsequent geometric transformation:
\begin{equation}\label{eq:fov_eq}
  \boldsymbol{M}_n(t,\theta,\phi)^i = \text{Rot}_{z}(\phi) \text{Rot}_{y}(\theta) \boldsymbol{M}_0^i + \boldsymbol{x}^{p}_n(t),\forall i \in\{1,..,5\}, \theta \in \Theta, \phi \in \Phi,
\end{equation}

\noindent where $\boldsymbol{M}_0 \in \mathbb{R}^{3\times5}$ denotes the FOV vertices of a downward facing camera centered at the origin of the 3D cartesian coordinate system and $\text{Rot}_{y}(\theta)$, $\text{Rot}_{z}(\phi)$ denote the 3D rotation matrices around the y-axis and z-axis respectively. Here, $\boldsymbol{M}_0^i$ represents the $i_\text{th}$ column of $\boldsymbol{M}_0$. Consequently, $\boldsymbol{M}_n(t,\theta,\phi)^i$ denotes the vertex of the FOV that has been rotated and translated accordingly. The matrices $\boldsymbol{M}_0$, $\text{Rot}_{y}(\theta)$ and $\text{Rot}_{z}(\phi)$ are respectively given by: 
\begin{equation}
    \begin{bmatrix}
       -\frac{c_l}{2} & \frac{c_l}{2} & \frac{c_l}{2}  & -\frac{c_l}{2} & 0 \\
        \frac{c_w}{2} &\frac{c_w}{2} & -\frac{c_w}{2} &  -\frac{c_w}{2} & 0 \\
        -c_r  & -c_r  &  -c_r  &  -c_r  & 0 \\
    \end{bmatrix},
    	\begin{bmatrix}
       \text{cos}(\theta) & 0 & \text{sin}(\theta)\\
       0 & 1 & 0\\
       -\text{sin}(\theta) & 0 & \text{cos}(\theta)    
    \end{bmatrix} \text{and}\begin{bmatrix}
       \text{cos}(\phi) & -\text{sin}(\phi) & 0\\
       \text{sin}(\phi) &  \text{cos}(\phi) & 0\\
        0 & 0 & 1
    \end{bmatrix}.    
\end{equation}

%In addition, we assume that the gimbal device is bounded to operate within a predefined finite set of admissible input rotation angles $\Xi = \Theta \times \Phi = \{(\theta,\phi) | \theta \in \Theta, \phi \in \Phi\}$, where $\times$ denotes the Cartesian product on the finite sets $\Theta$ and $\Phi$. Therefore, the camera FOV of each agent $j$ can take, at each time-step $k$, one out of $|\Xi|$ possible configurations ($|\Xi|$ denotes the cardinality of the set $\Xi$). 

Finally, we assume that at each time-step $t$ a finite number $N_R$ of (straight) light-rays modeling the direction of light propagation enters the camera's optical center and contribute to the imaging process. The set of light-rays captured by the agent's camera FOV $\boldsymbol{M}_n(t,\theta,\phi)$ is represented in this work by $\mathcal{R}_n(t,\theta,\phi) = \{\boldsymbol{R}_{n,1},\ldots,\boldsymbol{R}_{n,N_R}\}$ where $\boldsymbol{R}_{n,i}$ indicates an individual light-ray within the set. This light-ray is further characterized by the line segment:
\begin{equation} \label{eq:lightray}
    \boldsymbol{R}_{n,i} = \boldsymbol{r}_{n,i} + h[\boldsymbol{x}^{p}_n(t)-\boldsymbol{r}_{n,i}], \quad \forall i \in \{1,\ldots,N_R\}, h \in [0,1],
\end{equation}
where $\boldsymbol{x}^{p}_n(t)$ denotes the endpoint of the light-ray entering the camera's optical center at time-step $t$ (given by the agent's positional state) and $\boldsymbol{r}_{n,i} \in \mathbb{R}^3$ is a fixed point on the camera's FOV base that serves as the ray's origin. It is important to note that depending on the agent's positional state and camera rotation angles $\theta$ and $\phi$ a different set $\mathcal{R}_n(t,\theta,\phi)$ of light-rays occur as $\mathcal{R}_n(t,\theta,\phi)$ is a function of $\boldsymbol{x}^{p}_n(t)$, $\theta$ and $\phi$.

%Finally, it is assumed that at each time-step $k$ a finite set of (straight) light-rays, which model the direction of the propagation of light, enter the camera's optical center and cause matter to be imaged. The set of light-rays captured through the agent's camera FOV $\mathcal{V}^j_k(\theta^j_k,\phi^j_k,x_k^{j,\mathbf{p}})$ is denoted in this work as $\mathcal{L}^j_k(\theta^j_k,\phi^j_k,x_k^{j,\mathbf{p}}) = \{\Lambda^j_{k,1},..,\Lambda^j_{k,n}\}$, where $\Lambda^j_{k,i}$ denotes the individual light-ray in the set which is further given by the line-segment
%\begin{equation}\label{eq:lightray}
%    \Lambda^j_{k,i} = \lambda^j_{k,i} + d(x_k^{j,\mathbf{p}}-\lambda^j_{k,i}), ~\forall d \in [0,1],
%\end{equation}
%
%\noindent 
%where $x_k^{j,\mathbf{p}}$ is the light-ray's end point which enters the camera's optical center at time-step $k$, $\lambda^j_{k,i}$ is a fixed point on the camera's FOV base denoting the ray's origin, and $d$ is a scalar. Note here that every FOV state $\mathcal{V}^j_k(\theta^j_k,\phi^j_k,x_k^{j,\mathbf{p}})$ generates a different set of light-rays $\mathcal{L}^j_k(\theta^j_k,\phi^j_k,x_k^{j,\mathbf{p}})$.

\subsection{Collaborative Coverage Problem}\label{ssec:problem_statement}
The goal of the $N$ agents is to collaboratively cover (i.e., observe) with their cameras a given set $\mathcal{P} = \{\boldsymbol{p}_1,..,\boldsymbol{p}_{|\mathcal{P}|}\}, ~\boldsymbol{p}_{\tilde{p}} \in \mathbb{R}^3$ of points of interest ($\tilde{p}$ denotes the index of point $\boldsymbol{p}$) which reside on the surface area $\partial \mathcal{O}$ of an object of interest $\mathcal{O}$. Specifically, it is assumed that the object of interest $\mathcal{O}$ has been 3D reconstructed as a point-cloud and its surface area $\partial \mathcal{O}$ has been triangulated into a 3D triangle mesh $\mathcal{K}$ consisting of a finite set of non-overlapping triangular facets $\boldsymbol{\kappa} \in \mathcal{K}$ where $\boldsymbol{\kappa} \in \mathbb{R}^{3\times3}$ as shown in Fig. \ref{fig:fig0}(b). Consequently, our aim becomes the generation of collision-free collaborative coverage trajectories which cover all points $\mathcal{P}$ (where each point $\boldsymbol{p}_{\tilde{p}} \in \mathcal{P}$ resides on the corresponding facet $\boldsymbol{\kappa}_{\tilde{p}} \in \mathcal{K}$) on the object's surface area. Specifically, the problem tackled in this work can be stated as follows: \textit{
Given a team of $N$ agents $n = \{1,\ldots,N\}$, at each time-step $t$ design a set of collision-free collaborative look-ahead coverage plans (i.e., determine each agent's kinematic and camera control inputs) inside a rolling finite planning horizon $T$ which maximize the coverage of the points of interest $\mathcal{P}$ on the object's surface area.}

\section{Collaborative Rolling Horizon 3D Coverage Control}\label{sec:approach}

In order to tackle the coverage problem discussed above we formulate a distributed rolling horizon model predictive control (DMPC) problem, as detailed in Problem (P1). This controller seeks to find each agent's joint control inputs \(\{\boldsymbol{u}_n(t+\tau|t),~\theta_n(t+\tau|t),~\phi_n(t+\tau|t)\}, n \in \{1,\ldots,N\}\) over a rolling finite planning horizon \(\tau \in \{1,\ldots,T\}\) of length \(T\) time steps (i.e., predict the agents' states \(T\) time steps into the future), which optimize each agent's individual coverage objective function, denoted as \(\mathcal{J}_n\) and shown in Eq. \eqref{eq:objective_P1}, subject to a set of kinematic, sensing, and coverage planning constraints that aim to generate complementary coverage trajectories as detailed in Eq. \eqref{eq:P1_1} through Eq. \eqref{eq:P1_12}. In model predictive control, a longer planning horizon improves the optimality of results by considering a more extended future period but increases computational complexity, while a shorter horizon reduces computational demands at the expense of potentially suboptimal control actions.

\begin{algorithm}
\vspace{0mm}
\begin{subequations}
\begin{align} 
&\textbf{Problem (P1)}~\text{ -  Coverage Controller of Agent $n$:} & \notag\\
& \underset{\left\{\boldsymbol{u}_n(t+\tau|t),~\theta_n(t+\tau|t),~\phi_n(t+\tau|t)\right\}}{\arg \min} ~\mathcal{J}_n = \eta J_\text{guidance} + J_\text{coverage} &  \label{eq:objective_P1} \\
&\textbf{subject to: $\tau \in \{1,\ldots,T\}$} ~  &\nonumber\\
&\boldsymbol{x}_n(t+\tau|t) = A \boldsymbol{x}_n(t+\tau-1|t) + B \boldsymbol{u}_n(t+\tau|t) & \forall \tau\label{eq:P1_1}\\
&\boldsymbol{x}_n(t|t) = \boldsymbol{x}_n(t|t-1) &  \forall \tau\label{eq:P1_2}\\
&\boldsymbol{M}_{n,m}(t+\tau|t) =  \tilde{\boldsymbol{M}}_{n,m} + \boldsymbol{x}^p_n(t+\tau|t) & \forall \tau, m \label{eq:P1_3}\\
&\xi_{n,m,\tilde{p}}(t+\tau|t) = 1 \iff  p \in  \text{ConvHull} \left[\boldsymbol{M}_{n,m}(t+\tau|t)\right] & \forall \tau, m, \tilde{p} \label{eq:P1_4}\\
&\sum_{m=1}^{|\{\Theta \times \Phi\}|} \mu_{n,m}(t+\tau|t) = 1 &  \forall \tau, \label{eq:P1_5}\\
&\bar{\xi}_{n,m,\tilde{p}}(t+\tau|t) = \xi_{n,m,\tilde{p}}(t+\tau|t)~\wedge~\mu_{n,m}(t+\tau|t) & \forall \tau, m, \tilde{p}\label{eq:P1_6}\\
&\hat{\xi}_{n,m,\tilde{p}}(t+\tau|t) \leq \bar{\xi}_{n,m,\tilde{p}}(t+\tau|t) + \mathcal{Q}_n(\tilde{p}) +\sum_{k\ne n=1}^{N}\mathcal{Q}_k(\tilde{p}) & \forall \tau, m, \tilde{p} \label{eq:P1_7}\\
&\dot{\xi}_{n,m,\tilde{p}}(t+\tau|t)~\leq~ \hat{\xi}_{n,m,\tilde{p}}(t+\tau|t) ~+~ &  \label{eq:P1_8}\\
&~~~~~~~~~~~~\sum_{k<n} \sum_{m} \sum_{\tau} \dot{\xi}_{k,m,\tilde{p}}(t+\tau|t) + \sum_{k>n} \sum_{m} \sum_{\tau} \dot{\xi}_{k,m,\tilde{p}}(t-1+\tau|t-1)& \forall \tilde{p} \notag\\
&\boldsymbol{x}^p_n(t+\tau|t) \notin \text{ConvHull}[\mathcal{O}] & \forall \tau\label{eq:P1_9}\\
&\boldsymbol{x}_n(t+\tau|t) \in \mathcal{X},~ \boldsymbol{u}_n(t+\tau|t) \in \mathcal{U} & \label{eq:P1_10}  \\ 
&\xi_{n,m,\tilde{p}}(t+\tau|t), \bar{\xi}_{n,m,\tilde{p}}(t+\tau|t),\hat{\xi}_{n,m,\tilde{p}}(t+\tau|t),\dot{\xi}_{n,m,\tilde{p}}(t+\tau|t)  \in \{0,1\}   &   \label{eq:P1_11}\\ 
&\mu_{n,m}(t+\tau|t), \mathcal{Q}_n(\tilde{p}) \in \{0,1\}, \tilde{p} \in \{1,\ldots,|\mathcal{P}|\}, m \in \{1,\ldots,|\{\Theta \times \Phi\}|\}   & \label{eq:P1_12}
\end{align}
\end{subequations}
\vspace{0mm}
\end{algorithm}

In Problem (P1), at each time-step $t$ agent $n$ plans collaborative finite-length look-ahead coverage trajectories $\boldsymbol{x}_n(t+\tau|t), \tau \in \{1,\ldots,T\}$ which aim at optimizing the coverage of the points of interest $\mathcal{P}$ inside the planning horizon. The notation $\boldsymbol{x}_n(t^\prime|t)$ denotes the predicted agent state at time-step $t^\prime \geq t$ which was computed at time-step $t$. Therefore, the coverage planning problem is thus solved iteratively over multiple time-steps $t$ in a rolling horizon fashion where the first set of predicted control inputs in the sequence $\{\boldsymbol{u}_n(t+1|t),~\theta_n(t+1|t),~\phi_n(t+1|t)\}, n \in \{1,\ldots,N\}$ is executed in the next time-step the agents move to their new states and the optimization process shown above repeats for the next time-step over a shifted planning horizon until all points $\mathcal{P}$ are covered. 

Instead of addressing a large centralized optimization problem where all required information is sent to a central station that subsequently determines the control inputs for each agent as suggested in \cite{PapaioannouCDC2023} we break down the multi-agent coverage problem into smaller sub-problems that each agent can solve autonomously. Problem (P1) ensures that the control actions undertaken by one agent are consistent with those of all other agents in the team taking into account any interlinked coverage constraints between agents during the decision-making process. This is achieved through a coordination procedure \cite{Richards2007} where agent $n$ acquires the most recent plans from all preceding agents $i<n$ in the sequence and also gathers the projected plans from subsequent agents $i>n$ who have yet to finalize their latest plans.

In this approach the computational complexity of the proposed distributed coverage controller is decoupled from the number of collaborative agents, as opposed to a centralized formulation which becomes computationally intractable as the number of agents increases. In addition, we should point out that although this approach necessitates constant communication among the agents it enables the creation of complementary predictive look-ahead coverage plans that minimize the duplication of work. It is important to note that in certain application scenarios such as monitoring critical infrastructure structural inspection and area coverage for emergency response constant communication among the team of agents is crucial not only for operational efficiency but also for safety and security purposes. For example, in search-and-rescue operations once a victim is located, the discovering agent must quickly relay this information to the rest of the team which can then coordinate to provide assistance. Nevertheless, it is possible to relax the assumption of constant communication. In such cases, agents can opportunistically exchange plans whenever they are within communication range or apply the proposed coordination scheme within smaller neighborhoods. However, this flexibility comes at the cost of generating coverage plans that may exhibit overlaps and some duplication of work.
In Sec. \ref{ssec:distributed_control} we discuss in detail the proposed collaborative controller shown in Problem (P1). Subsequently, in Sec. \ref{ssec:light_path_constraints} we show how we extend this approach to include light-path propagation constraints to determine the visibility of the points of interest with respect to the agents' future predicted states, and we demonstrated how non-linear and non-convex visibility assessment constraints can been converted into logical expression which can easily be embedded into a mixed integer optimization framework.
 
% 
% proposed cooperative receding horizon 3D coverage controller formulated as mixed integer linear program (MILP) is shown in Problem (P2), which essentially produces the coverage trajectories i.e., the input reference trajectory of an auto-pilot system \cite{Simone2020,Elkaim2015,Garcia2011,Yang2016}, which in turn uses a low-level guidance controller (i.e., a flight controller) to execute the mission.

%which in this work is assumed to run on a mobile-base station which receives at each time-step the agents' states, and the facets they covered. 

%is shown in Problem (P2), formulated as a mixed integer linear program (MILP) for which the optimal solution can be obtained with existing optimization tools \cite{Anand2017}. Next we discuss the details of the proposed approach.

\subsection{Distributed Coverage Planning}\label{ssec:distributed_control}

Problem (P1) is executed by each agent $n$ in a coordinated manner to generate joint optimized coverage plans over a rolling planning horizon $\tau \in \{1,\ldots,T\}$ with the goal of covering points of interest in $\mathcal{P}$. As demonstrated, this objective is accomplished through a rolling horizon mixed integer program (MIP), the specifics of which are elaborated next.

\subsubsection{Collaborative Coverage Constraints} \label{sssec:constraints}
The constraints in Eq. \eqref{eq:P1_1}-\eqref{eq:P1_2} are due to the agent dynamics. These constraints are used to compute the predicted trajectory of agent $n$ inside the planning horizon by appropriately selecting the agent's direction and speed through the control input $\boldsymbol{u}_n(t+\tau|t), \forall \tau$. Subsequently, the constraint in Eq. \eqref{eq:P1_3} computes, all possible realisations of agent's $n$ camera states $\boldsymbol{M}_{n,m}(t+\tau|t)$ inside the planning horizon with respect to its predicted trajectoriy. Because the camera rotations involve non-linear transformations, as shown in Eq. \eqref{eq:fov_eq}, which make optimization challenging in this work we precompute all possible camera rotations at the origin. These are subsequently translated during optimization to the desired location based on the predicted trajectory. In other words, in order to embed camera rotation constraints into a linear MIP we precompute all possible camera states $\tilde{\boldsymbol{M}}_{n,m}$ at the origin as: $	\tilde{\boldsymbol{M}}_{n,m} = \text{Rot}_{z}(\phi) \text{Rot}_{y}(\theta) \boldsymbol{M}_0, \forall  m \in \{1,\ldots,|\{\Theta \times \Phi\}|\}$, where $|\{\Theta \times \Phi\}|$ is the cardinality of the set of angles which results from the Cartesian product between the finite sets $\Theta$ and $\Phi$. Subsequently, Eq. \eqref{eq:P1_3} translates for each agent all these possible camera states to its  predicted location $\boldsymbol{x}^p_n(t+\tau|t)$. Therefore, $\boldsymbol{M}_{n,m}(t+\tau|t)$ denotes the $m_\text{th}$ camera pose of agent $n$ at time-step $t+\tau|t$.

Next, for each point of interest $\boldsymbol{p}_{\tilde{p}} \in \mathcal{P}$ we associate a binary variable $\xi_{n,m,\tilde{p}}(t+\tau|t) \in \{0,1\}$ which indicates whether point $\boldsymbol{p}_{\tilde{p}}$ is covered by the $m_\text{th}$ camera state of agent $n$ at the future time-step $t+\tau|t$. This binary variable is activated (i.e., becomes equal to 1) when point $\boldsymbol{p}_{\tilde{p}}$ resides inside the convex hull of the $m_\text{th}$ camera FOV $\boldsymbol{M}_{n,m}(t+\tau|t)$ at time-step $t+\tau|t$ as shown in Eq. \eqref{eq:P1_4}.
A point $\boldsymbol{p}_{\tilde{p}} \in \mathcal{P}$ that belongs to the convex-hull defined by the camera FOV vertices in $\boldsymbol{M}_{n,m}(t+\tau|t)$ must satisfy the following system of linear inequalities: $\text{dot}(\boldsymbol{\alpha}_{i,n,m}(t+\tau|t), \boldsymbol{p}_{\tilde{p}}) \leq \beta_{i,n,m}(t+\tau|t), \forall i \in \{1,\ldots,5\}$, where $\text{dot}(\boldsymbol{a},\boldsymbol{b})$ is the dot product between vectors $\boldsymbol{a}$ and $\boldsymbol{b}$, $\text{dot}(\boldsymbol{\alpha}_{i,n,m}(t+\tau|t),\boldsymbol{p}_{\tilde{p}}) = \beta_{i,n,m}(t+\tau|t)$ is the equation of the plane which at time-step $t+\tau|t$ contains the $i_\text{th}$ face of the $m_\text{th}$ camera FOV state, of the $n_\text{th}$ agent $\boldsymbol{\alpha}_{i,n,m}(t+\tau|t)$ is the unit outward normal vector to this plane and $\beta_{i,n,m}(t+\tau|t)$ is a constant. Any point $\boldsymbol{p}_{\tilde{p}} \in \mathcal{E}$ which satisfies the aforementioned system of inequalities is contained within the convex-hull of $\boldsymbol{M}_{n,m}(t+\tau|t)$ and therefore can be potentially observed by the agent (provided it is visible). This functionality is then implemented as follows:
\begin{subequations}
\begin{align} 
&  \text{dot}(\boldsymbol{\alpha}_{i,n,m}(t+\tau|t),\boldsymbol{p}_{\tilde{p}}) + \tilde{\xi}_{i,n,m,\tilde{p}}(t+\tau|t)\left(M-\beta_{i,n,m}(t+\tau|t)\right) \le M,~ \forall i, m, \tau, \tilde{p} \label{eq:chull1}\\
& 5 \xi_{n,m,\tilde{p}}(t+\tau|t) - \sum_{i=1}^5 \tilde{\xi}_{i,n,m,\tilde{p}}(t+\tau|t) \le 0, ~ \forall  m, \tau, \tilde{p}. \label{eq:chull2}
\end{align}
\end{subequations}

\noindent In Eq. \eqref{eq:chull1} observe that the auxiliary binary variable $\tilde{\xi}_{i,n,m,\tilde{p}}(t+\tau|t) \in \{0,1\}$ becomes equal to 0 when the $i_\text{th}$ inequality is not satisfied (i.e., $\text{dot}(\boldsymbol{\alpha}_{i,n,m}(t+\tau|t),\boldsymbol{p}_{\tilde{p}}) \nleq \beta_{i,n,m}(t+\tau|t)$) and the constraints in Eq. \eqref{eq:chull1} holds with the use of a big positive constant $M$. Conversely, $\tilde{\xi}_{i,n,m,\tilde{p}}(t+\tau|t)=1$ when the $i_\text{th}$ inequality is satisfied. As a result the activation of $\xi_{n,m,\tilde{p}}(t+\tau|t) \in \{0,1\}$ for a particular configuration of the parameters $(m,\tilde{p},\tau)$ indicates that agent $n$ covers with the $m_\text{th}$ camera state point $\boldsymbol{p}_{\tilde{p}}$ at time-step $t+\tau|t$ which is achieved with the constraint shown in Eq. \eqref{eq:chull2}.

Next, the binary variable $\mu_{n,m}(t+\tau|t) \in \{0,1\}$ indicates which of the $|\Theta \times \Phi|$ camera states is active at time-step $t+\tau|t$. Essentially, $\boldsymbol{\mu}_n(t+\tau|t)$ is a matrix with $|\Theta \times \Phi|$ rows (indexed by $m$) and $T$ columns. At each time-step $t+\tau|t$ only one camera state should be active (i.e., the sum of each column should be equal to one) which is enforced via the constraint in Eq. \eqref{eq:P1_5}. The subsequent logical conjunction, i.e., Eq. \eqref{eq:P1_6} makes sure that only points that have been covered with the active camera state are considered as indicated by the binary variable $\bar{\xi}_{n,m,\tilde{p}}(t+\tau|t) \in \{0,1\}$.
Now, in order to handle the duplication of work (i.e., ensure that points of interest previously covered are not scheduled for future coverage) we need to keep track all points that have been covered. In a centralized formulation where the agents consult a central database $\mathcal{Q}(\tilde{p}) \in \{0,1\}, \forall \tilde{p} \in \{1,\ldots,|\mathcal{P}|\}$ which records the coverage status of each point $\boldsymbol{p}_{\tilde{p}} \in \mathcal{P}$ this can be accomplished with the following constraint: $\hat{\xi}_{n,m,\tilde{p}}(t+\tau|t) \leq \bar{\xi}_{n,m,\tilde{p}}(t+\tau|t) + \mathcal{Q}(\tilde{p}),~ \forall n, m, \tilde{p}$. For a point indexed by $\tilde{p}$ which has been already covered this constraint makes sure that the agents are discouraged from generating plans that include for coverage point $\boldsymbol{p}_{\tilde{p}}$ since the value of $\hat{\xi}_{n,m,\tilde{p}}(t+\tau|t)$ is activated through $\mathcal{Q}(\tilde{p})$ prompting the agents to focus on points yet to be covered. We will show in Sec. \ref{sssec:objective} how this binary variable can be linked to the coverage objective. 

However, in the absence of a central station the database $\mathcal{Q}(\tilde{p})$ is held in a distributed form amongst the agents i.e., $\mathcal{Q}(\tilde{p}) = \sum_{n=1}^N \mathcal{Q}_n(\tilde{p}), \forall \tilde{p}$ and therefore this functionality is implemented by allowing the agents to exchange at each time-step $t$ their individual records $\mathcal{Q}_n$ as shown in Eq. \eqref{eq:P1_7}.

To promote the generation of collaborative and complementary coverage plans that minimize work duplication the following constraint must be implemented: $\sum_{n} \sum_{m} \sum_{\tau} \hat{\xi}_{n,m,\tilde{p}}(t+\tau|t) \leq 1, ~\forall \tilde{p}$. This constraint ensures that during the planning horizon each point $\boldsymbol{p}_{\tilde{p}}$ is scheduled for coverage exactly once and by no more than one agent. In order to implement this constraint in a distributed fashion the agents generate their plans in a coordinated fashion and communicate their predicted plans to their peers. Specifically, agent $n$ receives the latest plans denoted as $\dot{\xi}_{k,m,\tilde{p}}(t+\tau|t), \forall k<n$ from all agents earlier in the sequence (i.e., for $k<n$) and the previous predicted plans from all agents (i.e., $k>n$) later in the sequence i.e., $\dot{\xi}_{k,m,\tilde{p}}(t-1+\tau|t-1) , \forall k>n$. Consequently, the generation of complimentary coverage plans is achieved with the constraint shown in Eq. \eqref{eq:P1_8}. The communication and information exchange protocol is further discussed in Sec. \ref{sec:approach}\ref{ssec:distributed_control}-\ref{sssec:communication}.

Finally, the constraint in Eq. \eqref{eq:P1_9} ensures that agent $n$ avoids collisions with the object of interest (and various obstacles in the environment). 
The convex-hull of the object of interest $\mathcal{O}$, is given by the intersection of $|\mathcal{K}|$ half-spaces (where $\mathcal{K}$ is the triangle mesh representing the surface area of $\mathcal{O}$). Suppose that the $i_\text{th}$ half-space is associated with the plane equation $\text{dot}(\boldsymbol{\alpha}_{i},\boldsymbol{x}) = \beta_{i},~ i\in \{1,\ldots,|\mathcal{K}|\}, \boldsymbol{x} \in \mathcal{E}$ which divides the 3D space into two parts. A collision with the object of interest can be avoided with the following constraints:
\begin{align}
&  \text{dot}(\boldsymbol{\alpha}_{i},\boldsymbol{x}^p_n(t+\tau|t)) + M o_{n,i}(t+\tau|t) > \beta_{i},~\forall \tau, i, \label{eq:O_1}\\
& \sum_{i=1}^{|\mathcal{K}|} o_{n,i}(t+\tau|t) \le  (|\mathcal{K}|-1), ~ \forall \tau \label{eq:O_2},
\end{align}
\noindent where $M$ is a big positive constant, and $o_{n,i}(t+\tau|t) \in \{0,1\}$ is a binary variable which on activation (i.e., Eq. \eqref{eq:O_1}) indicates that $\text{dot}(\boldsymbol{\alpha}_{i},\boldsymbol{x}^p_n(t+\tau|t)) \leq \beta_{i}$ holds for time-step $t+\tau|t$. Because agent $n$ is considered to collide with the object $\mathcal{O}$ at time-step $t+\tau|t$ (i.e., resides within the object's convex hull) when $\text{dot}(\boldsymbol{\alpha}_{i},\boldsymbol{x}^p_n(t+\tau|t)) \leq \beta_{i}$ for all $i$ the constraint in Eq. \eqref{eq:O_2} ensures that the number of times $o_{n,i}(t+\tau|t)$ is activated at each time-step during the planning horizon is less than or equal to $|\mathcal{K}|-1$. This condition is set to avoid collision. The constraints in Eq. \eqref{eq:O_1} - \eqref{eq:O_2} can be extended for multiple obstacles in the environment. 
The same principle can be applied for collision avoidance among the agents by making sure that the positional state of agent $n$ must remain outside the convex hull defined by a certain safety area $\mathcal{A}_{k}(t+\tau|t)$ of any other agent $k$ (where $k \neq n$ and $k \in \{1,\ldots,N\}$). For instance, $\mathcal{A}_{k}(t+\tau|t)$ could represent a convex area around $\boldsymbol{x}^p_{k}(t+\tau|t)$ serving as an approximation of a spherical safety zone with a specific radius around the agent as described in our previous work \cite{PapaioannouTMC2023}. Finally, the coverage mission concludes once every point $p$ has been covered a condition met when $\exists n \in \{1,\ldots,N\}: \sum_{\tilde{p}} \mathcal{Q}_n(\tilde{p}) = |\mathcal{P}|$.

\subsubsection{Agent Guidance and Coverage Objective} \label{sssec:objective}
Given the problem constraints discussed above and shown in Eq. \eqref{eq:P1_1} - \eqref{eq:P1_12} each agent $n$ can generate non-myopic coverage plans (i.e., a sequence of camera states $\boldsymbol{M}_{n,m}(t+\tau|t), m\in \{1,\ldots,|\{\Theta \times \Phi\}|\}$ along its predicted trajectory $\boldsymbol{x}_n(t+\tau|t)$) by minimizing the objective function $\mathcal{J}_n = \eta J_\text{guidance} + J_\text{coverage}$ which is composed of a guidance term as well as a coverage term defined respectively by $J_\text{guidance} = ||\boldsymbol{x}_n(t+1|t)-\boldsymbol{p}_n^\star||^2_2$ and $J_\text{coverage} = \sum_{\tau=1}^{T-1} \sum_{m=1}^{|\{\Theta \times \Phi\}|} \sum_{\tilde{p}=1}^{|\mathcal{P}|} \dot{\xi}_{n,m,\tilde{p}}(t+\tau+1|t)$. Therefore, the following objective function is minimized:
\begin{equation}\label{eq:objective_function}
	\mathcal{J}_n = \eta||\boldsymbol{x}^p_n(t+1|t)-\boldsymbol{p}_n^\star||^2_2 - \sum_{\tau=1}^{T-1} \sum_{m=1}^{|\{\Theta \times \Phi\}|} \sum_{\tilde{p}=1}^{|\mathcal{P}|} \dot{\xi}_{n,m,\tilde{p}}(t+\tau+1|t).
\end{equation}
\noindent where $\eta$ is a tuning weight. The first term (i.e., $J_\text{guidance}$) is a sub-objective responsible for guiding agent $n$ to its nearest unobserved point $\boldsymbol{p}_n^\star$. This ensures mission progress (avoiding deadlocks) especially in scenarios where no points of interest are reachable within the finite planning horizon. To determine this point at each time-step $t$ agent $n$ receives from all its peers their current positional states $\boldsymbol{x}^p_k(t|t), \forall k \neq n$, and computes $\boldsymbol{p}_n^\star$ by solving the following assignment problem:

\begin{equation} \label{eq:assignment}
	\hat{\boldsymbol{H}} = \min_{H} \sum_{k=1}^{N} \sum_{\tilde{p}=1}^{|\tilde{\mathcal{P}}|} \boldsymbol{C}(k,\tilde{p})\boldsymbol{H}(k,\tilde{p})
\end{equation}

\noindent where $k \in \{1,\ldots,N\}$ is the agent index, $|\tilde{\mathcal{P}}|$ represents the total number of unvisited points of interest obtained from the record $\mathcal{Q}$, $\boldsymbol{C}$ is a cost matrix containing the pairwise distances between the agents' positional states and the unobserved points of interest, and $\boldsymbol{H}(k,\tilde{p}) \in \{0,1\}$ is an assignment matrix with the property that each column and each row sum to one. Therefore, all agents $n \in \{1,\ldots,N\}$ solve at each time-step $t$ the same assignment problem shown in Eq. \eqref{eq:assignment} to jointly find their closest unobserved points of interest to target next, i.e., by determining the optimal assignment matrix $\hat{\boldsymbol{H}}$ and identifying $\boldsymbol{p}_n^\star$ from the $n_{\text{th}}$ row of $\hat{\boldsymbol{H}}$.

The second term ($J_\text{coverage}$) maximizes the number of points predicted for coverage within the planning horizon. The binary variable $\dot{\xi}_{n,m,\tilde{p}}(t+\tau|t)$ signals whether agent $n$'s predicted trajectory covers a point of interest ($\tilde{p}$) at a future time-step ($t+\tau|t$) with camera state $m$. Minimizing the negative of this variable optimizes the agent's movement and camera controls to increase coverage. 

%The optimization of this objective function also considers previously covered points and other agents' plans as detailed in Sec. \ref{sec:approach}\ref{ssec:distributed_control}-\ref{sssec:constraints}.

\subsubsection{Information Exchange} \label{sssec:communication}

To summarize, at each time-step $t$ each agent $n \in \{1,\ldots,N\}$ solves the distributed Model Predictive Control (MPC) problem outlined in Problem (P1) over a rolling finite planning horizon of length $T$ time-steps. The agents optimize their local objective function as described in Sec. \ref{sec:approach}\ref{ssec:distributed_control}-\ref{sssec:objective} with respect to their own control inputs and decision (binary) variables as detailed in equations \eqref{eq:P1_1} to \eqref{eq:P1_12} to generate complementary look-ahead coverage plans through coordination and information exchange. This process is facilitated by the following coordinated processing and information exchange protocol: At each time-step $t$ agent $n$ acquires the most recent plans from all preceding agents $k < n$ in the sequence $\dot{\xi}_{k,m,\tilde{p}}(t+\tau|t), \tau \in \{1,\ldots,T\}$ and also collects the previous predicted plans of all subsequent agents $k > n$ who have yet to generate their latest plans, $\dot{\xi}_{k,m,\tilde{p}}(t-1+\tau|t-1), \tau \in \{1,\ldots,T\}$. Because these binary decision variables indicate the points of interest $\tilde{p} \in \{1,\ldots,|\mathcal{P}|\}$ that are scheduled for coverage by other agents $k \neq n$ agent $n$ uses this information to generate plans that cover points not scheduled for observation by other agents. This is shown in Eq. \eqref{eq:P1_8} which essentially decentivizes agent $n$ for generating a coverage trajectory for a point of interest which  has been scheduled for coverage by preceding agents in the sequence while also accounting for points of interest that have been scheduled for coverage in the previous time-step by agents which have yet to generate their latest plans. Subsequently, the agents that generate plans after agent $n$ will take into account the latest plans of all previous agents $k \in \{1,\ldots,n\}$ and in turn they will revise their previous plans accordingly targeting points of interest not already scheduled for coverage.

Additionally, at each time-step $t$ agent $n$ receives from all other agents their record of already visited points, $\mathcal{Q}_k, \forall k \neq n$ which it combines to generate plans for unobserved points as captured by the binary variable $\hat{\xi}_n$ used in Eq. \eqref{eq:P1_8}.
Finally, agent $n$ also receives the positional state of all other agents $k \neq n$ at each time-step $t$ which is used for computing $\boldsymbol{p}^\star_n$ in the guidance objective as discussed in Sec. \ref{sssec:objective}. This information exchange protocol is repeated at each time-step enabling the agents to adapt their decisions on-line while considering the plans of their peers and minimizing duplication of work by generating complementary plans.

\subsection{Integrating Light-path Propagation Constraints}\label{ssec:light_path_constraints}
Although Problem (P1), generates non-myopic coverage plans which maximize the number of points of interest which reside inside the convex-hull of the agent's camera FOV inside the planning horizon it does not account for the notion of visibility as illustrated in Fig. \ref{fig:fig0}(c). In its current form Problem (P1) does not provide a way of determining which parts of the object's surface area are visible given the future planned state of the agent.

In the context of visibility determination consider the point $\boldsymbol{p}_{\tilde{p}}$ situated on a specific facet $\boldsymbol{\kappa}_{\tilde{p}}$ within the set $\mathcal{K}$. For $\boldsymbol{p}_{\tilde{p}}$ to be deemed visible at a future time-step $t+\tau|t$ through the field of view (FOV) of the camera $\boldsymbol{M}_{n,m}(t+\tau|t)$ it must reside within the convex hull of this FOV and subsequently visibility is affirmed if there is at least one light-ray $\boldsymbol{R}_{n,i}$ from the set $\mathcal{R}_{n,m}(t+\tau|t)$ for $i \in \{1,\ldots,N_R\}$ that  intersects lastly with $\boldsymbol{\kappa}_{\tilde{p}}$ which contains $\boldsymbol{p}_{\tilde{p}}$ before hitting the camera lens center. This set $\mathcal{R}_{n,m}(t+\tau|t)$ encapsulates all potential light-rays dictated by the camera's state $m$ at the given future time. To confirm visibility it must be ensured that the intersection operation symbolized by $\oplus$ between any light-ray $\boldsymbol{R}_{n,i}$ and the facets in $\mathcal{K}$ results in the facet $\boldsymbol{\kappa}_{\tilde{p}}$ which holds $\boldsymbol{p}_{\tilde{p}}$. If this intersection operation returns $\boldsymbol{\kappa}_{\tilde{p}}$ then $\boldsymbol{p}_{\tilde{p}}$ is considered visible from the camera state $\boldsymbol{M}_{n,m}(t+\tau|t)$. Conversely, if no intersections occur the operation results in $\emptyset$.

Define the equation of the plane containing the facet $\boldsymbol{\kappa}$ as $\text{dot}(\boldsymbol{\alpha}_{\kappa}, \boldsymbol{x}) = \beta_{\kappa}$ where $\boldsymbol{\alpha}_{\kappa} \in \mathbb{R}^3$ represents the unit normal vector perpendicular to the plane of $\boldsymbol{\kappa}$ and $\boldsymbol{x} \in \mathbb{R}^3$ denotes a point in space. The intersection point of the light-ray $\boldsymbol{R}_{n,i} = \boldsymbol{r}_{n,i} + h[\boldsymbol{x}^{p}_n(t)-\boldsymbol{r}_{n,i}]$, where $h \in [0,1]$, with the plane of facet $\boldsymbol{\kappa}$ is determined by the following set of equations:
\begin{subequations}
\begin{align}
    &\text{dot}\left(\boldsymbol{\alpha}_{\kappa}, \boldsymbol{r}_{n,i} + h[\boldsymbol{x}^{p}_n(t)-\boldsymbol{r}_{n,i}]\right) = \beta_{\kappa} \implies \label{eq:r1}\\
    &h = \frac{\beta_{\kappa} - \text{dot}(\boldsymbol{\alpha}_{\kappa}, \boldsymbol{r}_{n,i})}{\text{dot}(\boldsymbol{\alpha}_{\kappa}, \boldsymbol{x}^{p}_n(t)-\boldsymbol{r}_{n,i})} \label{eq:r2},
\end{align}
\end{subequations}
where Eq. \eqref{eq:r1} results from inserting the expression for $\boldsymbol{R}_{n,i}$ into the plane's equation and Eq. \eqref{eq:r2} solves for $h$. If the denominator in Eq. \eqref{eq:r2} equals zero it implies the light-ray is parallel to the facet leading to no intersection or an undefined one. Thus, for visibility the condition $\text{dot}\left(\boldsymbol{\alpha}_{\kappa}, \boldsymbol{x}^{p}_n(t)-\boldsymbol{r}_{n,i}\right) \neq 0$ is required to ensure a unique intersection. The intersection is valid if $h \in [0,1]$ and the intersection point $\hat{\boldsymbol{x}} = \boldsymbol{r}_{n,i} + \hat{h}[\boldsymbol{x}^{p}_n(t)-\boldsymbol{r}_{n,i}]$ (with $\hat{h}$ as the solution from Eq. \eqref{eq:r2}) lies within the convex hull of $\boldsymbol{\kappa}$ that is, $\hat{\boldsymbol{x}} \in \text{ConvHull}(\boldsymbol{\kappa})$.

The process described above must be applied at every time-step within the planning horizon to evaluate all potential realizations of light-rays $\boldsymbol{R}_{n,i} \in \mathcal{R}_{n,m}(t+\tau|t)$ determined by the agent's camera states $\boldsymbol{M}_{n,m}(t+\tau|t)$ and the triangular mesh $\mathcal{K}$ describing the object of interest. This assessment is crucial for determining the visibility of each point of interest $\boldsymbol{p}_{\tilde{p}} \in \mathcal{P}$ in relation to the agent's projected path. However, this method is computationally intensive and in addition requires the integration of non-convex and non-linear constraints (as illustrated in Eq. \eqref{eq:r2}) which complicate efficient optimization. To circumvent these difficulties this work adopts an alternative strategy that involves initially learning state-dependent light-path propagation constraints to assess visibility. These constraints are then integrated as logical constraints into the coverage controller outlined in Problem (P1).

This is achieved by first partitioning the environment $\mathcal{E}$ into a 3D grid $\mathcal{G}$ composed of discrete, non-overlapping cells denoted as $\mathcal{G}=\{\boldsymbol{G}_1,\dots,\boldsymbol{G}_{|\mathcal{G}|}\}$, with the union of all cells covering the entire grid $\bigcup_{g=1}^{|\mathcal{G}|} \boldsymbol{G}_g= \mathcal{G}$. Within each cell $\boldsymbol{G} \in \mathcal{G}$, $N_s$ joint configurations of the agent's position and camera state are randomly sampled, $\left(\boldsymbol{x}^{p}_{n,i} \in G, \boldsymbol{M}_{n,m,i}, m \in \{1,\ldots,|\{\Theta \times \Phi\}|\}\right)$ for each $i \in \{1,\ldots,N_s\}$. The visibility assessment outlined in Eq. \eqref{eq:r2} is then executed for these configurations to determine point visibility within each cell.
In essence for each cell  $g \in \{1,\ldots,|\mathcal{G}|\}$ we obtain a set of light-rays denoted as $\tilde{\mathcal{R}}_g = \bigcup_{i=1}^{N_s} \mathcal{R}_i$, where $\mathcal{R}_i$ corresponds to the set of light-rays obtained from the $i_\text{th}$ sample of the agent's positional and camera state in cell $\boldsymbol{G}_g$. Subsequently, we learn off-line the following state-dependant visibility assessment constraints for each point of interest and each cell: 
\begin{equation} \label{eq:vis_con1}
    v_{g,\tilde{p}} = 1 \iff \exists \boldsymbol{R} \in \tilde{\mathcal{R}}_g : \boldsymbol{R} \oplus \mathcal{K} = \boldsymbol{\kappa}_{\tilde{p}},~\forall g \in \{1,\ldots,|\mathcal{G}|\},\tilde{p} \in \{1,\ldots,|\mathcal{P}|\}
\end{equation}

\noindent Once the constraints above are learned we can determine the visibility of point $\boldsymbol{p}_{\tilde{p}}$ by utilizing the binary variable $v_{g,\tilde{p}}$ which is activated when there exists a light-ray $\boldsymbol{R}$ which traces back to facet $\boldsymbol{\kappa}_{\tilde{p}}$ when the agent resides within the cell $\boldsymbol{G}_g$. Subsequently, we integrate these visibility assessment constraints to the proposed coverage controller in Problem (P1) by extending Eq. \eqref{eq:P1_6} as follows:
\begin{equation} \label{eq:visibility_final}
	\bar{\xi}_{n,m,\tilde{p}}(t+\tau|t) = \xi_{n,m,\tilde{p}}(t+\tau|t)\wedge\mu_{n,m}(t+\tau|t)\wedge\left[v_{g,\tilde{p}} \wedge \varpi_{n,g}(t+\tau|t)\right], \forall m,\tilde{p},g,\tau,
\end{equation}
\noindent where the last part in the square brackets is due to the visibility assessment. Specifically, the binary variable $\varpi_{n,g}(t+\tau|t)=1 \iff \boldsymbol{x}^{p}_n(t+\tau|t) \in \text{ConvHull}[\mathcal{G}_g]$ is activated whenever agent $n$ resides within the convex-hull of cell $\mathcal{G}_g$, and subsequently this result is integrated with the visibility assessment binary variable $v_{g,\tilde{p}}$ which indicates whether the point $\boldsymbol{p}_{\tilde{p}}$ is visible from cell $\mathcal{G}_g$, further combined with the active FOV (i.e., $\mu_{n,m}(t+\tau|t)$), and the binary variable that indicates the coverage of point $\boldsymbol{p}_{\tilde{p}}$, i.e., $\xi_{n,m,\tilde{p}}(t+\tau|t)$, as shown in Eq. \eqref{eq:visibility_final}.

\section{Evaluation} \label{sec:eval}

\subsection{Experimental Setup} \label{ssec:sim_setup}

To evaluate the proposed approach we assume agents with identical capabilities are operating within a bounded 3D environment $\mathcal{E}$ which is confined in each dimension to the interval $[0,100]$m. The UAV agents are modelled based on the performance characteristics and specifications of the DJI Mavic Enterprise UAV platform. Subsequently, the agent's $n$ dynamics are characterised by $\Delta T=1$s, $\epsilon=0.3$, and $m=1.75$kg. The agent velocity is bounded within the interval $[-17,17]$m/s, whereas the control input $\boldsymbol{u}_n$ is bounded within the interval $[-10,10]$N. The tuning weight $\eta$ in Eq. \eqref{eq:objective_function} is set to 0.8. The agent camera FOV model parameters $(c_l, c_w, c_r)$ are set to $(8.5, 8.5, 10)$m, and the gimbal rotation angles $\theta$ and $\phi$ take their values from the finite sets $\Theta=\{30, 70, 110, 150\}$deg, and  $\Phi=\{30, 90, 150, 210, 270, 330\}$deg respectively. To determine visibility we have used $N_R=50$ light rays, $N_s$ was set to 100, and the environment was decomposed into $|\mathcal{G}|=100$ cells. Finally Problem (P1) was solved with the Gurobi solver running on a 3.2GHz desktop computer.

\begin{figure*}
	\centering
	\includegraphics[width=\textwidth]{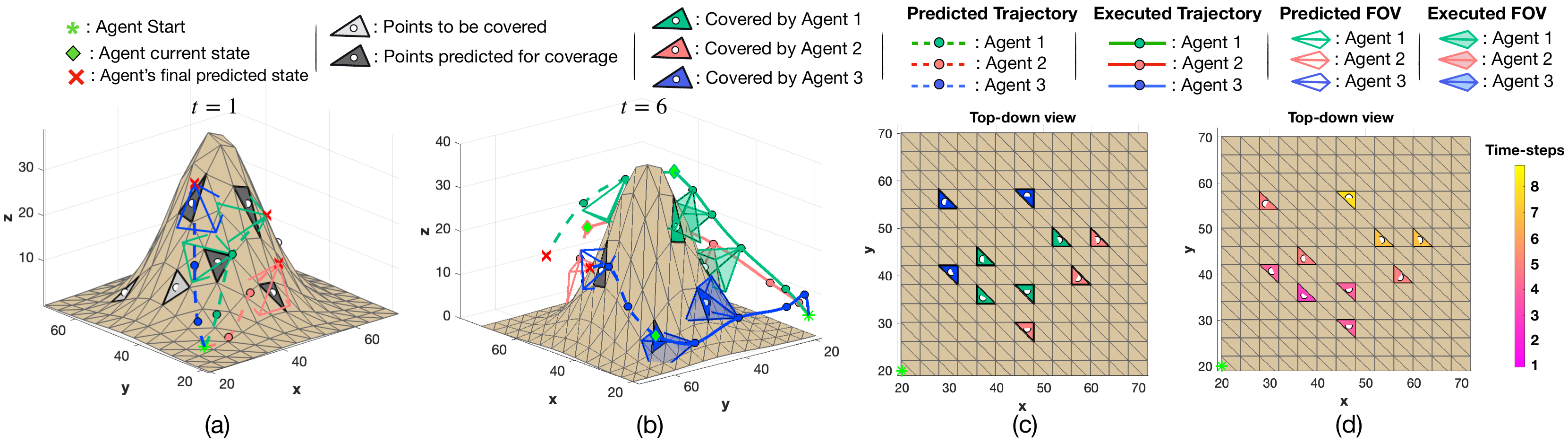}
	\caption{An illustrative example of the proposed collaborative 3D coverage controller, involving three agents indicated by green, red, and blue colors. (a) and (b) Predicted plans at time steps 1 and 6, respectively. (c) Allocation of points to the agents, and (d) the time step at which each point was covered.}
	\label{fig:res1}
	\vspace{0mm}
\end{figure*}

\subsection{Results}

An illustrative example of the proposed approach featuring 3 UAV agents is presented in Fig. \ref{fig:res1} for the coverage of 10 points on the surface area of the object depicted in the figure.
Specifically, Fig. \ref{fig:res1}(a) presents the predicted coverage plans (i.e., dotted lines) for the first time-step for 3 agents operating within a planning horizon of length $T=3$ time-steps all starting at the position $(x,y,z) = [20,20,5]$. This figure also illustrates the predicted camera Field of View (FOV) that optimizes coverage over the planning horizon. Triangular facets shaded in gray indicate the points of interest on these facets planned for coverage. The implemented coverage plans (i.e., the agents' trajectories and camera FOV states) are depicted with green, red, and blue lines for agents 1, 2, and 3 respectively. These colours also denote the points covered by each agent as shown in Fig. \ref{fig:res1}(b). As depicted in the figure the agents collaboratively plan non-myopic coverage strategies focusing on maximizing the coverage of uncovered points while minimizing the duplication of work. Finally, Fig. \ref{fig:res1}(c) offers a top-down view of how the points of interest have been allocated among the agents for coverage, and Fig. \ref{fig:res1}(d) indicates the point in time each point was covered.

\begin{figure*}
	\centering
	\includegraphics[width=\textwidth]{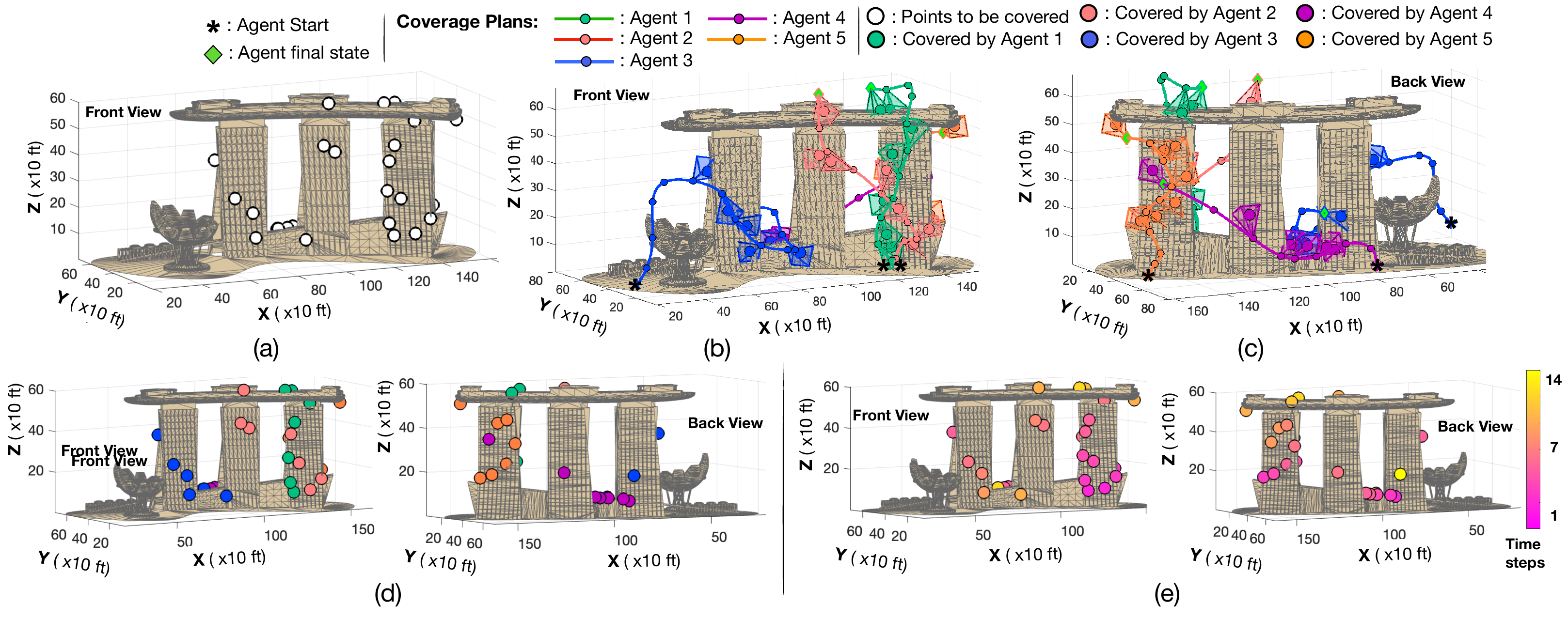}
	\caption{Coverage of the Marina Bay Sands Hotel in Singapore with 5 collaborative UAV agents. (a) Points of interest for coverage shown as $\circ$,  (b)(c) Final coverage trajectories, (d) Allocation of points to agents, and (e) Timing of point coverage during the mission. }
	\label{fig:res2}
	\vspace{0mm}
\end{figure*}

Next, Fig. \ref{fig:res2} shows the proposed approach applied for the task of 3D structural inspection of a real-world building i.e., to a 3D reconstruction of the Marina Bay Sands (MBS) hotel in Singapore. The MBS, consisting of three towers as seen in Fig. \ref{fig:res2}(a), stands approximately 200m (656 ft) tall. These towers are linked by a 340m (1120 ft) long skybridge as depicted. In this experiment we randomly selected 35 points (marked as white $\circ$) from the triangulated surface mesh of the building as illustrated in Fig. \ref{fig:res2}(a). Five agents are positioned at the coordinates $(125,10,5)$, $(115,10,5)$, $(20,20,5)$, $(80,80,5)$, and $(160,60,5)$ for agents 1 (green), 2 (pink), 3 (blue), 4 (purple), and 5 (orange) respectively. Maintaining the previously discussed setup, Fig. \ref{fig:res2}(b)(c) display the final executed coverage plans by the five agents achieving full coverage of all points of interest within 15 time-steps. The proposed approach generated collaborative trajectories for optimized coverage with the allocation of points to agents clearly shown in Fig. \ref{fig:res2}(d) and the timing of point coverage during the mission shown in Fig. \ref{fig:res2}(e). We should note here that, in order to handle obstacle avoidance in such complex, non-convex structures, the object of interest is first decomposed into several convex sub-regions. This decomposition allows us to utilize the constraints shown in Eq. \eqref{eq:P1_9} in each of those sub-regions to avoid collisions.

\begin{figure*}
	\centering
	\includegraphics[width=\textwidth]{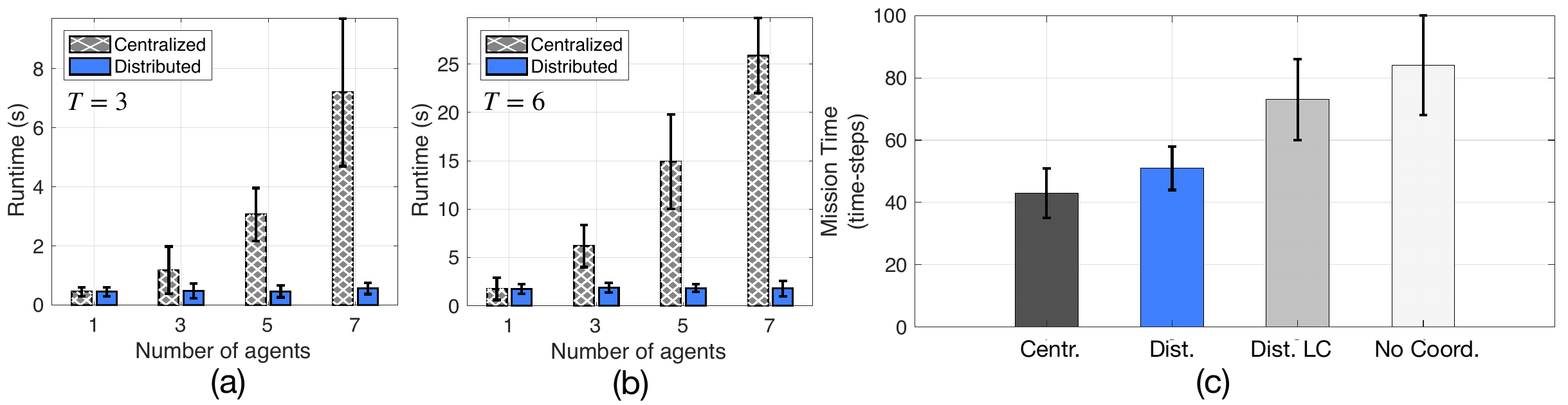}
	\caption{Performance evaluation of the proposed approach. (a)(b) Computational complexity of centralized (\cite{PapaioannouCDC2023}) and distributed (proposed approach) coverage planning with respect to the number of agents and the length of the planning horizon. (c) Performance comparison with competing approaches.}
	\label{fig:res3}
	\vspace{0mm}
\end{figure*}

Our next goal is to evaluate the proposed approach's performance concerning computational complexity as shown in Fig. \ref{fig:res3}(a)(b), and performance i.e., mission completion time, as depicted in Fig. \ref{fig:res3}(c), comparing it with other methods. For this purpose, we have conducted a Monte Carlo (MC) simulation randomly initializing the states of $N$ agents, where $N \in \{1,3,5,7\}$, within the environment $\mathcal{E}$ as per the setup described in Sec. \ref{sec:eval}\ref{ssec:sim_setup} for covering 35 points of interest on the object depicted in Fig. \ref{fig:res1}. The first set of experiments measures the average runtime required to solve one iteration of Problem (P1) and generate a solution, varying the planning horizon's length and the number of agents, and then comparing these findings with the centralized approach from \cite{PapaioannouCDC2023}. It is important to mention that the computational complexity of a mixed integer program (MIP), such as Problem (P1), is generally influenced by the number of binary variables. This is because the primary optimization method for solving MIPs, a branch-and-bound variant \cite{vielma2015mixed,morrison2016branch}, constructs a search tree that enumerates potential solutions systematically. The size of this tree which directly affects computational complexity is determined by the number of integer and binary variables.
Therefore, the primary factor affecting computational complexity, as illustrated in Problem (P1), is the length of the planning horizon. Conversely, the centralized approach mentioned in \cite{PapaioannouCDC2023} also depends on the number of agents for generating collaborative coverage plans. Figure \ref{fig:res3}(a) presents the average runtime over 50 MC trials for generating coverage plans with our distributed approach for a planning horizon of $T=3$ time-steps (blue bar) compared to the centralized approach (gray hatched bar) for scenarios with 1, 3, 5 and 7 agents. As evident from Fig. \ref{fig:res3}(a), while both approaches are comparable in runtime for the single agent scenario, the centralized controller's computational complexity increases exponentially with the number of agents indicating poor scalability. Conversely, the proposed distributed approach's performance is not influenced by the number of agents, maintaining its ability to generate non-myopic collaborative plans efficiently. Similar findings are depicted in Fig. \ref{fig:res3}(b) for a planning horizon of $T=6$, where the proposed method's performance is influenced only by the planning horizon's length, not by the number of agents.

The subsequent experiment assesses the performance of the proposed approach in terms of mission completion time, i.e., the duration required for agents to cover all points of interest. This analysis involved 50 MC trials with $N=5$ agents randomly placed within the environment to cover 35 points of interest, using a planning horizon of length $T=6$. Besides comparing to the centralized approach we also evaluate two variations: distributed with limited communication (Dist. LC) and multi-agent coverage without coordination (No Coord.). In the Dist. LC setup, agents collaborate opportunistically within a 3m communication range with the lowest-index agent initiating the proposed approach. Agents within communication range also exchange their coverage databases $Q_n$, whereas those outside generate plans independently following Problem (P1). The No Coord. approach involves agents always planning independently without collaborative strategies, though agents within communication range share their coverage databases $\mathcal{Q}_n$. 

It is important to mention here that with regard to the approaches Dist. LC and No Coord., in scenarios where the agents are not in communication range, the guidance objective in Eq. \eqref{eq:objective_function} cannot be computed due to the lack of information required for solving the assignment problem. For this reason, each agent $n$ computes its nearest unobserved point of interest in a greedy manner by calculating: $\arg\min_{\tilde{p}} \|\boldsymbol{x}_n(t+\tau|t) - \tilde{\mathcal{P}}_n(\tilde{p})\|^2_2$, where $\tilde{\mathcal{P}}_n$ is a list of unobserved points of interest based solely on agent $n$'s record of observed points $\mathcal{Q}_n$.

Figure \ref{fig:res3}(c) displays the average mission completion time for these four methods. The centralized approach completes the mission in about 40 time-steps, with the proposed distributed method being roughly $15\%$ slower, but 6.5 times more computationally efficient, as shown in Fig. \ref{fig:res3}(b). The Dist. LC variant highlights the significant role of communication in forming collaborative plans and influencing mission completion times, further emphasized by the No Coord. approach's performance which lacks inter-agent communication for collaborative plan generation. These experiments illustrate the trade-offs between computational complexity, performance, and communication, showcasing the proposed approach's adaptability to diverse scenarios and needs.

\begin{figure*}
	\centering
	\includegraphics[width=\textwidth]{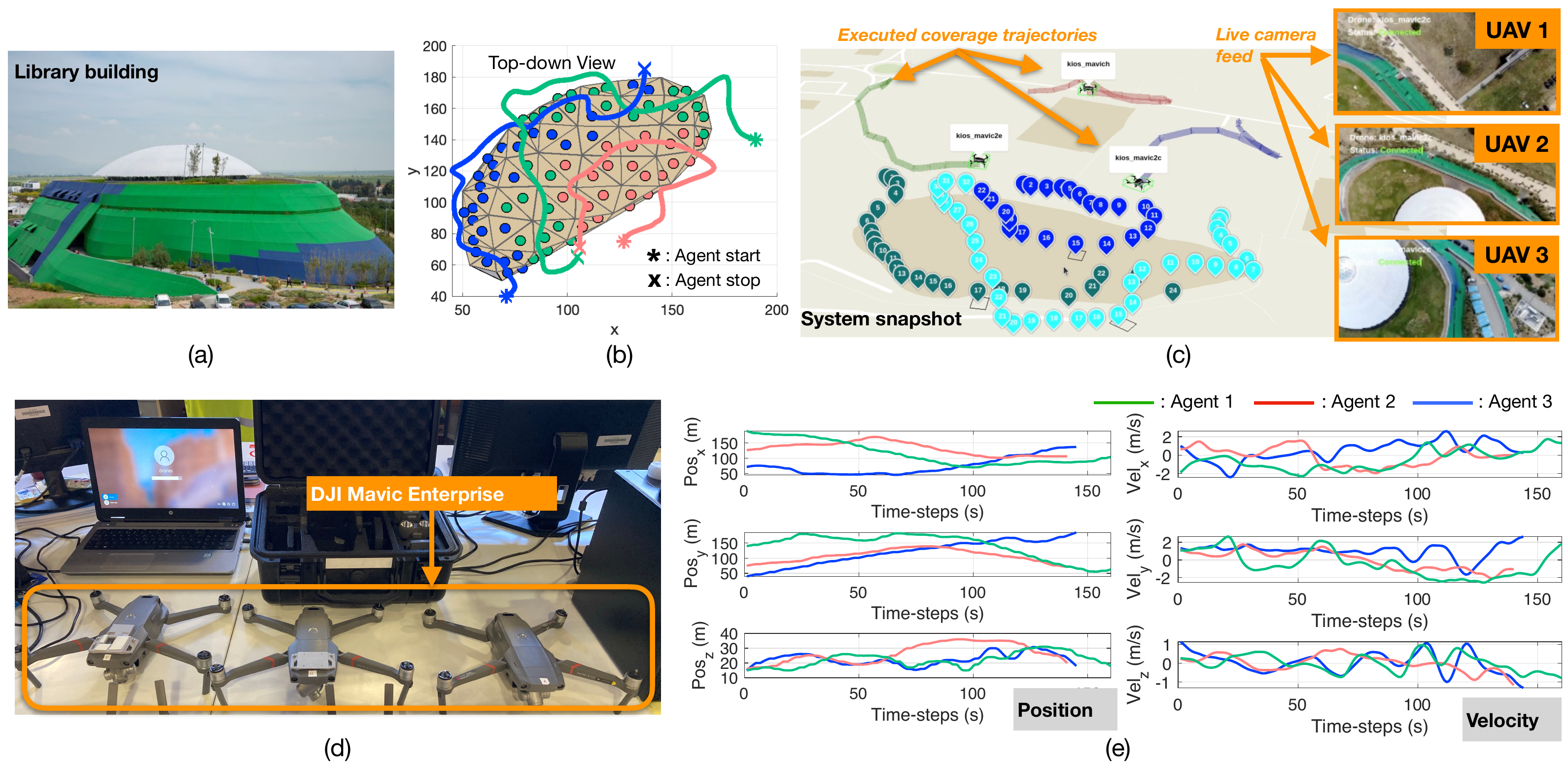}
	\caption{Real-world experimental evaluation of the proposed collaborative 3D coverage controller: (a) Building for coverage, (b) Generated collaborative coverage plans with 3 UAV agents, (c) Real-time monitoring of mission progress, (d) DJI Mavic drones used in the experiment, (e) Telemetry data.}
	\label{fig:res4}
	\vspace{-5mm}
\end{figure*}

Finally, we have conducted experimental tests of the proposed approach for covering the surface area of the University of Cyprus Library building, an oval-like structure with major and minor axis diameters of approximately 170m and 85m, respectively, and a height of about 25m, as depicted in Fig. \ref{fig:res4}(a). The building's surface was first converted from a 3D point cloud into a mesh of 82 triangular facets as illustrated in Fig. \ref{fig:res4}(b). Using the proposed approach we generated collaborative coverage trajectories for 3 UAV agents tasked with covering the building's surface area, depicted in Fig. \ref{fig:res4}(b). This figure specifically presents a top-down view of the building and illustrates the trajectories generated by the 3 agents, ensuring coverage of the entire area. The covered points are colour-coded to match the colour of the agents that covered them i.e., green, red, and blue, for agents 1,2 and 3 respectively.

Implementation details: The UCY library building area was pre-scanned using a single quadrotor UAV equipped with a camera and lidar. This process involved capturing multiple images and raw lidar data, which were subsequently processed to generate a 3D point cloud. The point cloud was triangulated to create the 3D mesh shown in Fig. \ref{fig:res4}(b).
The experimental setup utilized three DJI Mavic Enterprise UAVs, as illustrated in Fig. \ref{fig:res4}(d). Each UAV was equipped with a 12MP gimbal-stabilized camera and operated at a safety-maximizing velocity of 2 m/s. The UAVs were controlled using the DJI Mobile Software Development Kit (SDK), which enabled the development of a custom Android mobile application running on mobile phones assigned to the UAVs. Each mobile phone running the custom SDK-enabled application wirelessly connected to its respective drone's remote controller, acting as an intermediary to facilitate command transmission to the drone and the reception of telemetry data (i.e., Fig. \ref{fig:res4}(e)), such as GPS location, altitude, battery status, speed, and sensor information.
Due to SDK constraints and hardware limitations (specifically, the lack of direct access to the drones' onboard controllers), the proposed distributed model predictive coverage controller could not be implemented on the UAVs' onboard control systems. Instead, a ground control station (GCS) executed the distributed controller in Matlab and transmitted the resulting trajectories to the UAVs through the mobile applications running on the Android phones connected to each drone's remote controller. The GCS also received telemetry data from the UAVs via this application. Communication between the GCS and the UAVs was established through a VPN server over a wireless network, allowing for data exchange between agents and the GCS. The live mission monitoring was conducted using our dedicated multi-UAV multi-tasking platform \cite{Terzi2019}, as illustrated in Fig. \ref{fig:res4}(c).
While this prototype implementation was not optimal, it successfully demonstrated the proposed approach's effectiveness and revealed limitations and areas for future improvement. Notably, reliance on GPS positioning introduced errors and inconsistencies in UAV localization, resulting in distorted coverage plans in certain cases. Additionally, communication delays due to network latency affected the UAVs' operation, and environmental disturbances caused deviations from the planned trajectories. 
Finally, with the current prototype implementation, we were unable to evaluate the real-time capabilities of the proposed controller or thoroughly assess its performance in real-world scenarios. For future work, we aim to co-design the software and hardware components of the proposed controller to facilitate its deployment on real-world platforms. Promising avenues for further research include the development of efficient real-time mixed-integer programming (MIP) controllers via adaptive neighborhood search techniques \cite{Hendel2021}, and learning-based approaches \cite{wu2021learning}.
\color{black}

\section{Way Forward in UAV Swarm Systems}\label{sec:wf}

Swarm systems have the potential to revolutionize society, the economy, technology, and various other fields by enhancing efficiency, scalability, and adaptability across a wide range of applications. In emergency response, swarms of autonomous drones and ground vehicles can cooperatively patrol \cite{patrolling} and search areas not previously encountered, locate victims, gather real-time information, and execute complex missions with minimal human oversight, thereby enhancing situational awareness and operational capabilities. Meanwhile, in military contexts, swarm technologies have the potential to create coordinated defense mechanisms for surveillance and reconnaissance. These fields exemplify the potential of swarm system technologies in tackling complex challenges across diverse sectors through the application of collective intelligence principles. However, despite recent progress in various aspects of swarming systems \cite{zhou2020uav, zhu2016recent, Soon-Jo2018}, numerous challenges remain before these systems reach the required level of maturity. These challenges range from technological issues, such as the development of smarter, more compact machines, to interactive aspects, including the creation of innovative interfaces and modes of communication, and extend to socio-technical issues related to evaluating trustworthiness and addressing the ethical and legal ramifications of deploying swarm systems. In this work, we address several limitations of existing multi-UAV and swarm systems for coverage planning by developing an intelligent distributed model predictive controller. This controller incorporates light-path propagation constraints, allowing for the generation of optimal, complementary look-ahead trajectories for the efficient coverage of 3D objects of interest. However, numerous open challenges and limitations persist, and overcoming these obstacles will be crucial for advancing coverage planning and the field of swarm systems in general, as elaborated next: 

a) Scalability and Coordination: A major challenge for UAV swarm systems is their scalability. As the swarm size grows, managing communication, coordination, and control becomes exponentially more complex. Existing algorithms and frameworks, including this work, struggle to scale effectively when the number and interactions among the UAVs increases. Currently this problem is handled by various distributed frameworks which although can handle large number of agents, they only generate myopic and greedy plans \cite{Soon-Jo2018}. On the other hand, to enable the generation of complimentary look-ahead coverage plans, this work requires a coordinated optimization sequence to be performed among the agents, which can potentially limit its applicability with large teams. 
Another limitation of the proposed technique is the dependence of the results on the agents' order of execution. This issue arises from the way the centralized problem was broken down and approximated with a distributed formulation. Essentially, by evaluating all possible permutations of the agents' execution order, the centralized solution can be recovered. However, this is computationally intractable, which is why we proposed this distributed approach. Future research should explore whether, in certain scenarios, there is an opportunity to adopt an intelligent strategy to determine the optimal sequence of agent execution, and prioritize the development of algorithms that are inherently scalable and capable of handling large numbers of UAVs without sacrificing performance. There are several avenues to explore here including learning-based techniques \cite{bertsimas2022online}, as well as planning without coordination by allowing some degree of work duplication as shown in our previous work \cite{PapaioannouTSMC}.

b) Autonomy and Decision Making: A crucial aspect of future UAV swarms is their ability to operate autonomously in complex and unknown environments while making intelligent and optimal decisions. This involves not only navigating through unfamiliar terrain but also determining the most efficient way to accomplish a given task while rapidly adapting to new information and changing conditions. One of the limitations of the proposed approach is the requirement of a known world model, which is essential for designing optimal coverage plans using the current methodology. Specifically, the proposed method requires prior knowledge of the 3D environment and points of interest, as well as pre-computed light-path propagation constraints. However, this information may not be available in many scenarios where UAV-based search is needed, such as search and rescue in natural disaster zones or inside collapsed buildings. Currently, state-of-the-art swarm systems are divided between model-based and model-free planning methodologies \cite{zhou2020uav}. Model-based methodologies \cite{zhu2016recent} rely on predefined environmental models for optimized planning, offering efficiency but lacking adaptability to changes. In contrast, model-free techniques like sampling-based approaches \cite{Jones2023} adapt to unknown or dynamic environments, providing flexibility but lacking optimality. The proposed methodology can be extended to handle dynamic and unknown environments by employing a two-stage approach: first, building a world model through environmental mapping and scene reconstruction, and then utilizing this model for planning. This two-stage approach has been explored in our previous work \cite{AnastasiouICUAS2024} with a swarm of heterogeneous agents, demonstrating promising results. However, to enhance swarm system performance, future research should focus on creating more integrated or hierarchical planning techniques that can synergistically leverage the benefits of both model-based and model-free approaches. Additionally, integrating machine learning techniques \cite{westheider2023multi} and other artificial intelligence methodologies \cite{booch2021thinking,papaioannouIJCNN} can enhance the swarm's ability to learn from experience, improve its performance over time, and handle complex decision-making.

c) Robustness and Fault Tolerance: Robustness with respect to modelling errors and measurement noise, as well as fault tolerance with respect to hardware malfunctions is paramount for UAV swarm systems.. To operate effectively in real-world environments, swarms must be resilient to unpredictable conditions, varying weather, and hardware malfunctions. Future research should focus on developing advanced robust and fault-tolerant algorithms that enable the swarm to dynamically adapt to UAV losses, malfunctions, and uncertainty. Additionally, incorporating machine learning techniques could allow UAVs to predict potential failures and proactively adjust their behavior to prevent mission-critical consequences. While this work did not study robustness and fault tolerance, our previous research on fault-tolerant coverage planning under non-Gaussian disturbances \cite{PapaioannouIFAC} can provide valuable insights for future directions. Additionally, recent advancements in these fields, as highlighted in \cite{ftc3}, offer further potential avenues for exploration. 

Swarm systems, with their potential to revolutionize various fields, offer a promising solution to complex challenges that traditional centralized systems struggle to address. While significant strides have been made, numerous challenges remain to fully harness their potential. By addressing the challenges discussed above, swarm systems can become indispensable tools for solving a wide range of real-world problems.

\section{Conclusion} \label{sec:conclusion}
This study introduces a coverage control framework that allows a group of distributed autonomous agents to collaboratively plan look-ahead coverage plans over a rolling finite planning horizon, targeting the coverage of designated areas on a 3D object's surface. Formulated as a distributed model predictive control problem, the proposed approach optimizes the movement and visual control inputs of each agent, incorporating constraints to minimize overlapping work among agents. By integrating visibility determination through light-path propagation constraints, the controller can anticipate which areas of the object will be visible based on projected future positions of the agents. This is achieved by transforming non-linear visibility assessments constraints into logical constraints using binary variables within a mixed-integer programming framework. The effectiveness of this approach is validated both in simulation environments and through real-world UAV inspections of architectural structures.

\section{Acknowledgements}
This work is supported by the European Union's Horizon 2020 research and innovation programme under grant agreement No 739551 (KIOS CoE), and from the Government of the Republic of Cyprus through the Cyprus Deputy Ministry of Research, Innovation and Digital Policy.

\flushbottom
\balance

\bibliographystyle{IEEEtran}
\bibliography{main} 

\end{document}